# GDN: A Stacking Network Used for Skin Cancer Diagnosis

Jingmin Wei[a], Haoyang Shen[b], Ziyi Wang[c], Ziqian Zhang[d]

[a]School of Artificial Intelligence and automation, Huazhong University of Science and Technology, Wuhan, Hubei, China 430074, jmwei@hust.edu.cn;
[b]Biosystem Engineering, Zhejiang University, Hangzhou, Zhejiang, China 310000, 3180100265@zju.edu.cn;
[c]Computer Science Department, New York University, New York, USA 10012, zw1718@nyu.edu;
[d]School of Remote Sensing and Information Engineering, Wuhan University, Wuhan, Hubei, China 430079, zhangziqian@whu.edu.cn;

## ABSTRACT

Skin cancer, the primary type of cancer that can be identified by visual recognition, requires an automatic identification system that can accurately classify different types of lesions. This paper presents GoogLe-Dense Network (GDN), which is an image-classification model to identify two types of skin cancer, Basal Cell Carcinoma, and Melanoma. GDN uses stacking of different networks to enhance the model performance. Specifically, GDN consists of two sequential levels in its structure. The first level performs basic classification tasks accomplished by GoogLeNet and DenseNet, which are trained in parallel to enhance efficiency. To avoid low accuracy and long training time, the second level takes the output of the GoogLeNet and DenseNet as the input for a logistic regression model. We compare our method with four baseline networks including ResNet, VGGNet, DenseNet, and GoogLeNet on the dataset, in which GoogLeNet and DenseNet significantly outperform ResNet and VGGNet. In the second level, different stacking methods such as perceptron, logistic regression, SVM, decision trees and K-neighbor are studied in which Logistic Regression shows the best prediction result among all. The results prove that GDN, compared to a single network structure, has higher accuracy in optimizing skin cancer detection.

## 1. INTRODUCTION

Skin cancer has become the most common type of cancer in the United States and worldwide in present days according to the Skin Cancer Foundation[1]. Fortunately, the symptoms of skin lesions are usually visible by human eyes and skin cancer can be easily controlled if found from an early stage. Skin cancer has five stages regarding its severity, with stage-0 indicating that the cancer cells are contained only in the top layer of skin and stage-4 indicating the most severe case that melanoma has spread elsewhere in the body, away from where it started and the nearby lymph nodes[2]. Research conducted by Cancer Research UK has shown that almost 100% of patients would survive their cancer for 5 years or more after they are diagnosed if they are in stage-1[3]. Therefore, early diagnosis of skin cancer can be promising to achieve more efficient skin cancer treatment to increase the survival rate.

To diagnose skin cancer for early treatment, the availability of proper screening methods to detect and classify the initial symptom of different skin lesions is required. A variety of machine learning techniques have been used for skin cancer detection in recent years due to their capability of computing large quantity of information from image input. Among them, deep neural networks have shown their unique strength by reaching higher prediction accuracy with multiple-layer structures. Some heavily-applied convolutional neural networks include ResNet[4], VGGNet[5], DenseNet[6], GoogLeNet-Inception[7], etc. ResNet presents a residual learning framework to ease the training of networks while keeping the depth of the computation. VGGNet features with its utilization of small filters because of fewer parameters and stacking more of them instead of having larger filters. For DenseNet, the layers are densely connected, with each layer gets the input from previous layers output feature maps. GoogLeNet allows a balance between increasing the depth and width of the network while keeping the computational budget constant at the same time. However, most researches conducted so far mainly focus on the continuous refinement of one particular network, which results in either relatively low prediction accuracy or highly expensive computation cost with a large number of iterations over a large dataset.

†These authors contributed equally.

In this paper, we propose a new model, the GoogLe-Dense Network (GDN), which uses the combined prediction of GoogLeNet and DenseNet to diagnose whether and what kind of skin cancer a patient has. Firstly, we preprocess the ISIC images (an open source skin cancer image dataset on Kaggle) by normalized clipping, format conversion and random augmentation on valid data while deleting duplicate and useless data. Then, we apply GoogLeNet and DenseNet to construct and train two separate models that can detect skin cancer. Next, to improve the accuracy of the method, we apply the ensemble learning strategy called stacking[8] to synthesize models from the previous step. In this stacking step, the classifier we use is logistic regression[9]. Finally, we get our synthetic model GDN (GoogLe-Dense Network) for skin cancer recognition.

To verify the effectiveness of our method, we compare our model with the other four baselines on the same ISIC-images dataset, including VGGNet, GoogLeNet, ResNet and DenseNet. The prediction accuracy of GDN on the skin disease dataset is 82.2%, and the prediction accuracy of other CNN models are 69.9% (GoogLeNet), 76.2% (DenseNet), 32.9% (ResNet121) and 32.4% (VGG16). We also compare the accuracy of different weak classifiers used in stacking besides logistic regression, including perceptron[10], support vector machine[11], K-neighbor[12] and decision tree[13]. The precision of Logistic regression is 82.2%, better than other weak classifiers. The accuracy rate of other weak classifiers are 78.7% (perceptron), 70.4% (SVM), 74.6% (K-neighbor) and 71.8%(decision tree). The experimental results and analysis show that GDN, compared to a single network structure, demonstrates its higher accuracy in solving skin cancer detection problem.

## 2. THE PROPOSED GOOLE-DENSE NETWORK(GDN)

This section introduces our GDN model. As shown in figure 1, our model mainly contains two consequential levels. To construct the GDN model, we first preprocess the data to fit the network input and improve the training efficiency (Sec. 2.1). Then, we train GoogLeNet and DenseNet respectively with the preprocessed dataset (Sec. 2.2) and (Sec. 2.3). Finally, we use the stacking method to complete the GDN model for skin cancer detection (Sec. 2.4).

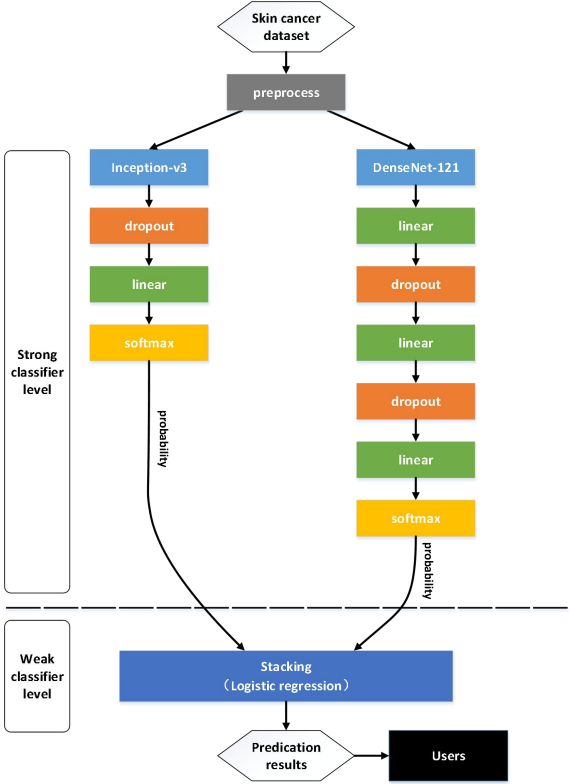

Figure 1. The Structure of the GDN model

## 2.1 Data Preprocessing

The dataset contains 6 types of skin lesion images in total, with 2 positive categories (skin cancer types) and 4 negative categories. Since duplicated images are considered a case of nonrandom sampling, they are deleted first when the dataset is loaded to avoid model overfitting in the later training. After deleting duplicates, each image in the dataset is resized to 256×256 pixel, followed by a random 224×224 cropping and a random horizontal flipping. These are the data augmentation techniques we use to improve the robustness of the model. To standardize the input, these images are also normalized to tensors ranging from 0 to 1. For our training, we split the data with 80% images being the training set and the remaining 20% the validation set. This results in 2339 training samples and 602 validation samples for our model training. The data output after preprocessing is shown in figure 2.

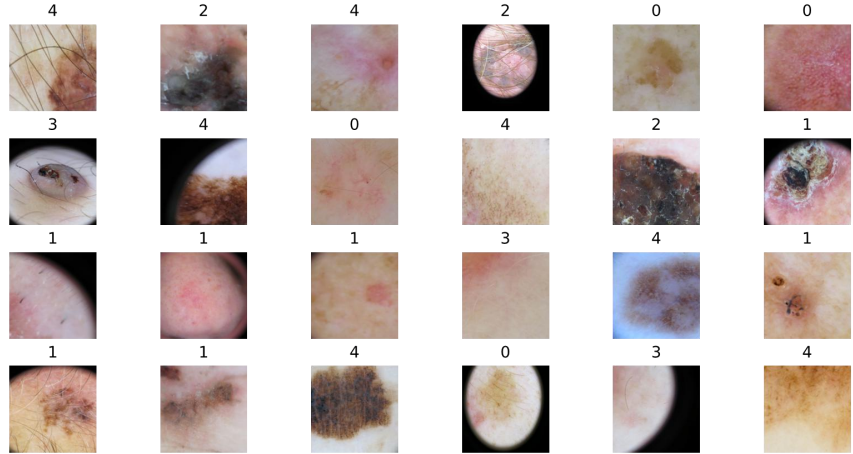

Figure 2. One batch training data after preprocessing

Table 1. GoogLeNet Model

| Type | Patch Size/stride or remarks | Input Size |
|---|---|---|
| Inception-v3 | googlenet model | 268203 |
| dropout | P=0.5 | 2048 |
| linear | 2048-6 | 2048 |
| softmax | classifier | 6 |

Table 2. DenseNet Model

| Type | Patch Size/stride or remarks | Input Size |
|---|---|---|
| DenseNet-121 | densenet model | 150528 |
| linear | 25088-512 | 25088 |
| dropout | P=0.5 | 512 |
| linear | 512-256 | 512 |
| dropout | P=0.2 | 256 |
| linear | 256-6 | 256 |
| softmax | classifier | 6 |

## 2.2 Model Construction

GoogLeNet[7] can increase the depth and breadth of the network on the premise that computing power requirements remain unchanged. And inception block of GoogLeNet is a structure with parallel connections. As shown in figure 1, we build Inception-v3[14], add a dropout layer (0.5) and adjust the full connection layer (2048-6) to fit 6 category prediction. Besides, in the previous data enhancement step, we also change the cropping size from 224×224 to 299×299 to accommodate the input of the model.

DenseNet[6] follows the idea of residual block for optimization. It increases the depth of the network and reduces the number of parameters. It proposes a dense block, which is a feature-multiplexed structure. We build Densenet121[6] network and optimize its structure. At the end of the network, we add a full connection-ReLU (FCR for short) layer (25088-512), a dropout layer (0.5), a FCR layer (512-256), a dropout layer (0.2), a FCR layer (256-6), and finally a softmax layer. In this way, the network can realize 6-category classifier and largely prevent overfitting at the same time.

## 2.3 Model Training

In the training process of neural network, Adam optimizer[15] is used to update the gradient. The main principle of Adam is to consider first moment estimation and second moment estimation of gradient and calculate updating step of each iteration. Besides, we use cross entropy loss to calculate the loss. After setting the learning rate and the number of iterations, the training process begins. The iterative method divides each process into two steps (figue 3):

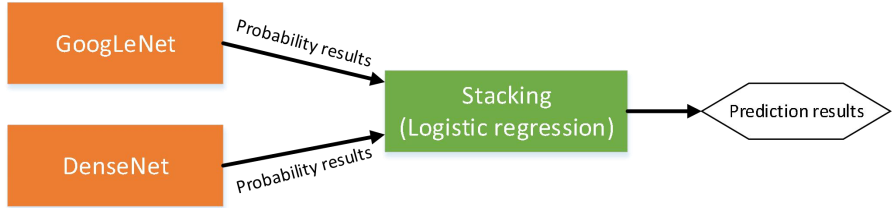

Figure 3. Model stacking

- Update parameter of the network on the training set.
- Compute performance of the network on the validation set.

In Step a, we first calculate and save the predicted value of the network on the training dataset. Then we calculate and save the loss based on cross entropy[16] equation (1). After that, we empty the past gradient and back propagate the loss, and finally update the network parameters according to the gradient.

$$loss(x, class) = -\log \frac{\exp(x[class])}{\sum_i x[i]} = -x[class] + \log(\sum_i \exp(x[i])) \qquad (1)$$

After training, we use saved predicted values and labels of the training dataset to calculate the number of correct predictions and divide it by the total number of the training dataset to calculate precision based on equation (2).

$$accuracy = \frac{correct\ prediction\ numbers}{total\ numbers}. \qquad (2)$$

In Step b, we eliminate the step of emptying the past gradient, back propagating the loss and updating the network parameters. The other steps are the same as Step a.

In the training and validation process, we apply three commonly used methods to solve the over-fitting and under-fitting problems. Firstly, as mentioned above, we appropriately change model structure such as adding dropout layers to effectively prevent overfitting. Besides, according to the training accuracy, we moderately modify data augmentation. Furthermore, based on the training accuracy visualization, we choose whether to stop or continue the training process in advance.

We define every 50 epochs as one training round. According to the performance of each 50 epochs in the training and validation dataset, we determine whether to continue training. For DenseNet which has a relatively complex structure, over-fitting may occur during training. To improve the robustness of the network, our solution is to add dropout layers and more complex data augmentation. We also stop the training process in advance when the network has an over-fitting tendency based on accuracy visualization. For GoogLeNet which has a relatively simple structure, under-fitting may occur during training. Thus, we add more rounds (50 epochs per) during training until accuracy reaches more than 70%.

Finally, we save the probability of network output and two networks for the convenience of building the GDN model. In the training process, we also improve the training efficiency by parallel training of the two networks.

In the training process of stacking, we are using Logistic regression[9] with L1 regularization. The main idea is to derive the Sigmoid function based on the maximum likelihood and the Bernoulli distribution assumption of the sample and use this function as a classification model.

$$h(X_i) = \theta(X_i^T w) = \frac{1}{1+\exp(-X_i^T w)} \tag{3}$$

According to relevant literature[19], the optimization principle of Logistic regression with L1 regularization is as shown in equation (4).

$$\min_{w,c} ||w||_1 + C\sum_{i=1}^{n} \log(\exp(-y_1(X_i^T w + c)) + 1) \tag{4}$$

In equation (4), C is a constant used to control the intensity of L1 regularization. The model input of the second-level classifier is the probability of the two networks of the first-level classifier, and the output is the final classification result. In the model training process, we use cross-entropy loss, gradient descent optimization algorithm, and L1 regularization to optimize according to the training set data and the corresponding label. In the model verification process, we also calculate the accuracy of the model on the verification set according to equation (2), and visualize the relevant results.

In the end, we got a stacking network GDN.

## 3. EXPERIMENTAL SETTINGS

This section describes the dataset (Sec. 3.1), related baselines (Sec. 3.2), evaluation metrics (Sec. 3.3) and implementation details (Sec. 3.4) in this paper.

### 3.1 Dataset

We use the open source skin lesions dermatoscopic image dataset from Skin Lesions-Dermatoscopic Images[17], which is originally downloaded from International Skin Imaging Collaboration (ISIC). The International Skin Imaging Collaboration (ISIC) is an academic partnership. The main purpose of this partnership is to promote the application of skin imaging, promote cooperation between dermatology and computer science, and develop better diagnostic methods. The ISIC file is the largest public data set in the field of dermatoscopic image analysis and research. It contains more than 60,000 images. This data set has a unified classification standard and is recognized by experts in the medical field. At present, many researchers in the medical and scientific fields use this data. Collected products. The data set we use has a smaller amount of data, which is suitable for model building and training. This dataset contains 6 categories of skin lesions in total, which includes Actinic Keratoses, Basal Cell Carcinoma, Seborrheic Keratoses / Solar Lentigo, Dermatofibroma lesions, Melanoma, and Vascular lesions. Melanoma is a type of skin cancer that develops from the pigment-producing cells known as melanocytes. Basal Cell Carcinoma is a type of cancer that often appears as a slightly transparent bump on the skin. These are the two types of cancer contained in the dataset. All the other categories are noncancerous skin growth, except vascular lesions which can be either benign or malignant. Still, all these remaining categories are treated as negative samples that are noncancerous in our experiment.

### 3.2 Baselines

VGG has proposed the idea of constructing a deep model by reusing simple basic blocks. A comprehensive evaluation of the ever-increasing depth of the network using an architecture with very small (3×3) convolutional filters shows that by increasing the depth to 16-19 weight layers, significant improvements to existing technology configurations can be achieved.

GoogLeNet is a network with parallel connections. It increases the depth and breadth of the network on the premise that the computing power requirements remain unchanged. GoogLeNet uses a modular structure (Inception structure) for easy addition and modification. At the same time, GoogLeNet draws on the ideas in NiN[18], adds convolution, and optimizes the Inception block, which improves the accuracy of the model. To avoid the disappearance of the gradient, GoogleNet additionally adds an auxiliary classifier to forward the gradient.

ResNet has proposed a residual learning framework to ease the training of networks, which are deeper than previously used networks. It explicitly reconstructs the layer to learn the residual function about the input of the layer, instead of learning the unreferenced function. It also provides comprehensive empirical evidence that the residual network is easy to optimize and can increase the accuracy by significantly increasing the depth.

DenseNet follows the idea of residual blocks for optimization, and has proposed densely connected blocks: For each layer, the feature maps of all previous layers are used as input, and its feature map is used as the input of all subsequent layers. The network alleviates the vanishing gradient problem, strengthens feature propagation, encourages feature reuse, and greatly reduces the number of parameters.

### 3.3 Evaluation Metrics

For separate network training, the training of GoogLeNet and DenseNet use the same evaluation metrics to measure its performance. For each iteration, both the training and validation set apply cross-entropy loss based on equation (1) and divide it by the number of all samples to calculate the loss. To calculate the accuracy is to divide the number of correct samples by the number of all samples. After the iterative process, the accuracy of the output model is calculated by dividing the number of correct samples by the number of all samples in the validation set based on equation (2). For the stacking part, the accuracy is measured the same way as the model accuracy.

### 3.4 Implementation Details

In the step of data preprocessing, we split the origin images with 80% being the training set and 20% the validation set. Each image is resized to 256×256, followed by a random 224×224 cropping (299×299 in GoogLeNet) and a random horizontal flipping. Then we define mean value=[0.485,0.456,0.406], standard deviation=[0.229,0.224,0.225] for data normalization. In the step of constructing the data loader, we set batch size=24 and shuffle the dataset in each epoch in GoogLeNet, set batch size=8 and shuffle the dataset in each epoch in DenseNet. For the network design in GoogLeNet, we build Inception-v3, add a dropout layer (p=0.5) and alter the last fully connected layer (2048-6). For the network design in DenseNet, we optimize DenseNet121 by adding several fully connected layers at the end of the network. Sequentially, they are full connection-ReLU layer (25088-512), dropout layer (p=0.5), full connection-ReLU layer (512-256), dropout layer (p=0.5), full connection-ReLU layer (256-6) and softmax layer. To train the models, we use Adam as the optimizer and set the learning rate to 0.001 in GoogLeNet and 0.01 in DenseNet. Cross entropy loss is the loss function in both models. We define every 50 epochs as one training round. GoogLeNet and DenseNet are trained in 3 and 4 rounds simultaneously. Besides, in the second level of GDN, the parameters of logistic regression are as follows: C=0.02126(regularization strength), penalty term=L1, solver=saga(this solver support non-smooth L1), tol=0.1(criteria to stop solving).

## 4. EXPERIMENTAL RESULTS

In this section, we discuss the overall performance of our model, compared with other baselines (Sec. 4.1). A detailed analysis of GDN and our insight is also provided (Sec. 4.2).

### 4.1 Overall Performance

At present, there are many useful CNN models which have successfully addressed many issues in different fields. However, in skin cancer diagnosis, due to the diversity of patients' disease conditions, many CNNs perform difficulty in detection, both in accuracy and speed. Our model has shown some improvements in respect to these issues. Based on the fusion method of GoogLeNet and DenseNet, we have achieved a higher recognition accuracy rate than other models.

We compared the results of the GDN model with the existing CNN model, and the results show that GDN has a better performance in skin disease recognition. The prediction accuracy of GDN on the skin disease dataset is 78.3%, and the prediction accuracy of other CNN models is GoogLeNet (69.9%), DenseNet (76.2%), ResNet (32.9%) and VGG (32.4%).

## 4.2 Training Process

In the strong classifier level, GoogLeNet iterates a total of 300 epochs, and DenseNet iterates a total of 250 epochs. At the same time, we test the accuracy after training. As shown in figure 4 and figure 5, the accuracy of the two networks exceeded 70%.

In the following weak classifier level, we use logistic regression to stack the two networks. Besides, the output of this logistic regression model is used as the final output of our GDN. We test our model on the skin disease data set, and the final prediction accuracy rate is 82.2%, proving that our model is better than various existing models.

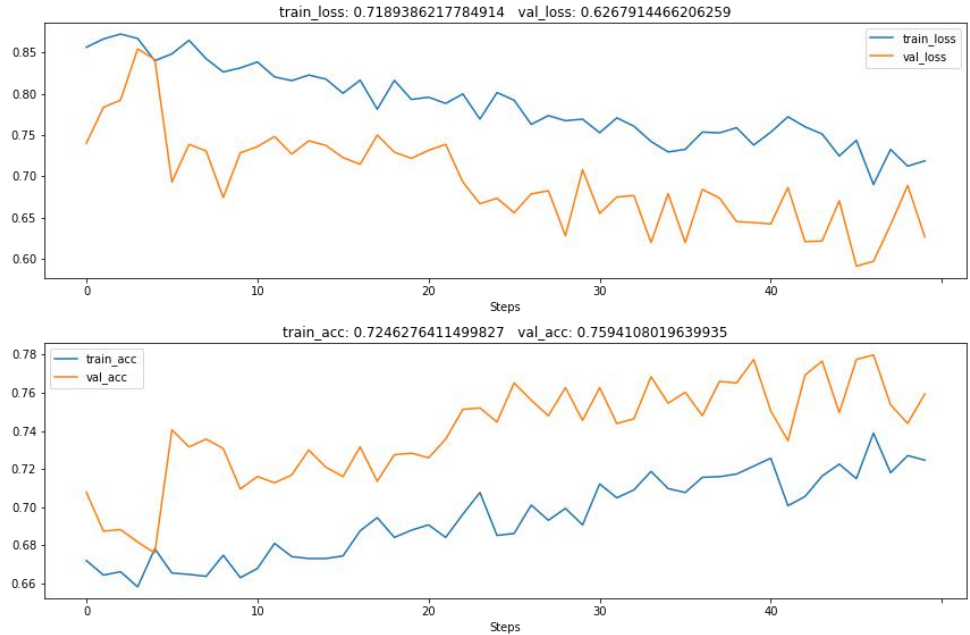

Figure 4. The training process of GoogLeNet (the last 50 epochs)

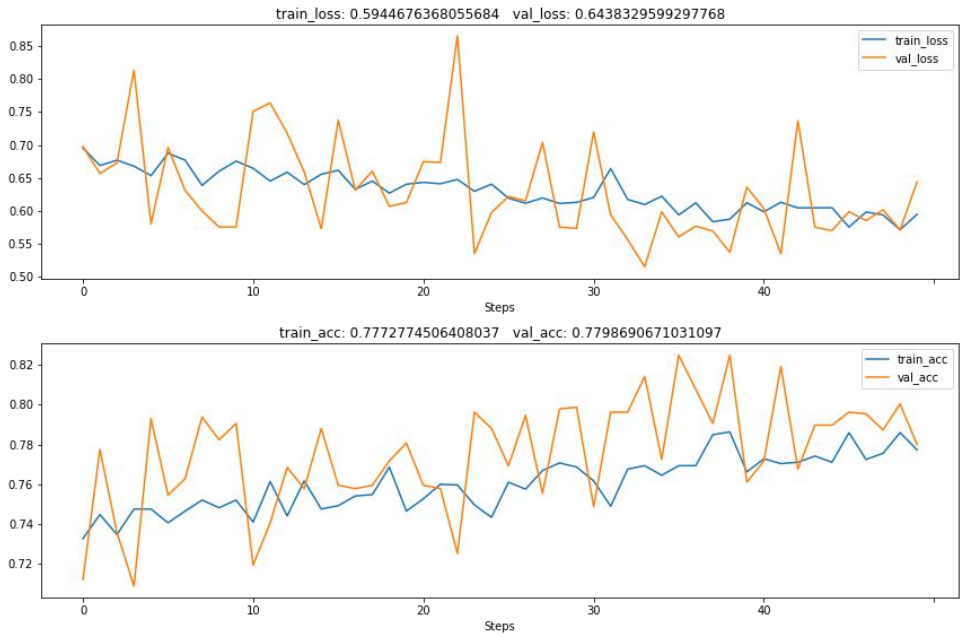

Figure 5. The training process of DenseNet (the last 50 epochs)

### 4.3 Ablation Study

Logistic regression is a type of generalized linear regression analysis model, which is often used in data mining, automatic disease diagnosis, economic forecasting, and other fields. It is better applied to multi-classification problems, and the training speed is faster. For the diagnosis of skin cancer, logistic regression can be well applied to our scenarios. Compared with perceptron, logistic regression can converge well to linear inseparable data. Compared with SVM, logistic regression is more sensitive to extreme values, and the training efficiency is higher. Compared with the K-neighbor algorithm, the logistic regression model is not easy to be misled by a small part of the error data. Compared with the decision tree algorithm, logistic regression is more suitable for the overall analysis of the data. Logistic regression is essentially the use of maximum likelihood to estimate the posterior probability distribution of the sample, and the use of cross-entropy to vote and make decisions. To a certain extent, it can more comprehensively consider different characteristics and reduce misclassification and misjudgment.

Therefore, in the weak classifier level, we choose a logistic regression model. In the diagnosis of skin cancer, its classification accuracy is better than other weak classifiers. As shown in figure 6, its classification accuracy rate is 82.2%. The accuracy rate of other weak classifiers is 78.7% (perceptron), 70.4% SVM), 74.6% (K-neighbor) and 71.8% (decision tree).

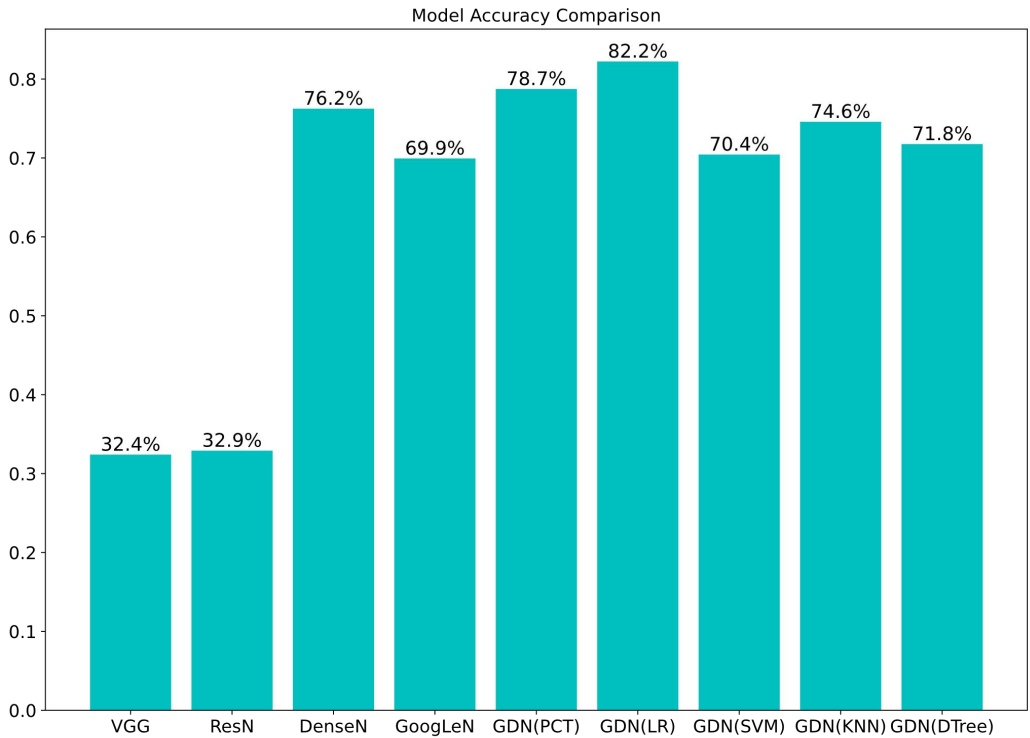

Figure 6. Accuracy comparison among different models

### 4.4 Discussion

The diagnosis of skin diseases has always been a difficult problem. Due to the different skin colors, different lesion locations, and the similarities between different types of skin lesions, the current various CNN networks are not effective in the diagnosis of skin cancer. However, the GDN network combines the advantages of GoogLeNet and DenseNet and performs diagnosis, which has a certain improvement in accuracy. This will help patients understand their skin diseases more quickly and take appropriate measures. At the same time, it can also help doctors quickly make a diagnosis and improve their efficiency.

# 5. CONCLUSION

In this paper, we propose the GDN model, which is a stacking network, to do skin cancer detection with both high accuracy and high efficiency. Our main point in the research is to first train GoogLeNet and DenseNet in parallel and then take the two trained models as the input for a logistic regression model to get the final prediction result.

To confirm the strength of our network, some other models are also tested and our experimental results show that GDN, compared to a single network structure, has proved its efficiency and accuracy in skin cancer prediction. To put the GDN model for practical use, we provide a simple user interface as in figure 7. In this application, we have implemented image preprocessing, image transformation, GDN model prediction, image format conversion, prediction result output, input format error checking and user warnings, and other functions. Based on our GDN model, we have built this skin disease recognition system completely. For users who use the application, they only need to upload skin images, and the system will recognize the images and give possible skin disease types, possibilities, and visual results. As shown in figure 7, we also give diagnostic suggestions according to the different prediction diagnoses.

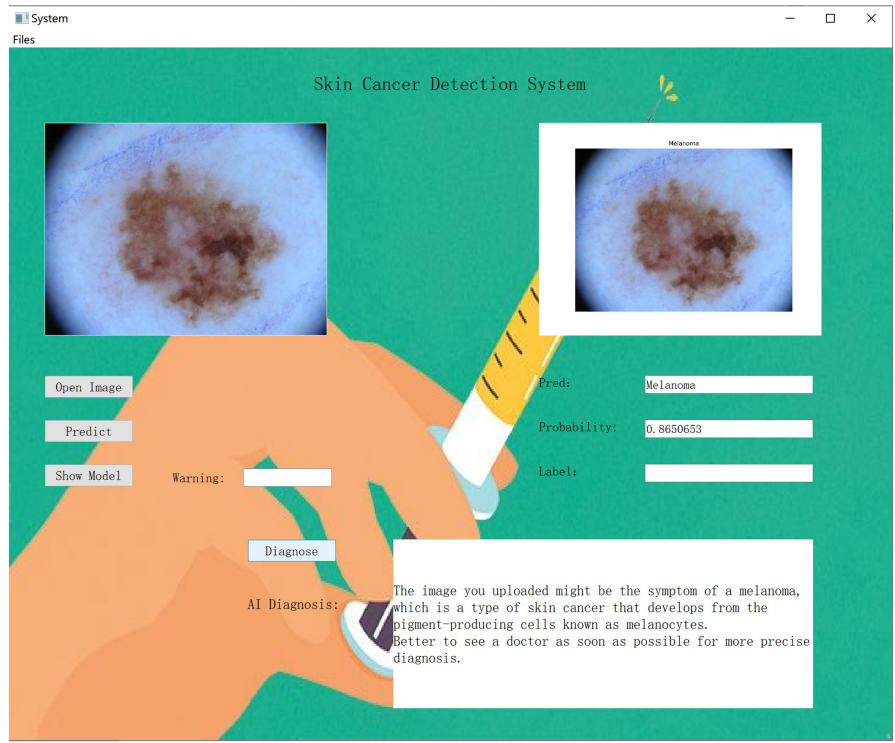

Figure 7. User's interface of GDN model

In the future, our study will focus on the expansion of GDN into the detection of other cancer types that are often diagnosed. With the fact that early diagnosis benefits the health of the patients and diagnosis through trained machine model saves the use of limited medical resources in hospital, we hope that GDN could be better improved in the future to achieve better performance and put into use practical.